\documentclass[runningheads]{llncs}
\usepackage{booktabs}    
\usepackage{graphicx}
\usepackage[marginal]{footmisc}
\usepackage{amssymb}
\usepackage{amsmath} 
\usepackage{caption}
\usepackage{lipsum}
\usepackage{multirow}
\usepackage{array}
\usepackage{hhline}
\usepackage{marvosym} 
\renewcommand{\thefootnote}{}
\newcommand{\tabincell}[2]{\begin{tabular}{@{}#1@{}}#2\end{tabular}}

\begin{document}
\title{KESDT: knowledge enhanced shallow and deep Transformer for detecting adverse drug reactions}
\titlerunning{KESDT for detecting adverse drug reactions}
%

\author{Yunzhi Qiu$^{1}$ \and Xiaokun Zhang$^{1}$ \and Weiwei Wang$^{1}$ \and Tongxuan Zhang$^{2}$ \and Bo Xu$^{1}$ \and Hongfei Lin$^{1}$$^{(\textrm{\Letter})}$}
\authorrunning{Y. Qiu et al.}
%
\institute{$^{1}$School of Computer Science and Technology, Dalian University of Technology, China\\
$^{2}$College of Computer and Information Engineering, Tianjin Normal University, China\\
\email{yzqiu@mail.dlut.edu.cn}, 
\email{hflin@dlut.edu.cn}}

\maketitle              

\sloppy  

\renewcommand{\thefootnote}{\arabic{footnote}} 

\begin{abstract}
Adverse drug reaction (ADR) detection is an essential task in the medical field, as ADRs have a gravely detrimental impact on patients' health and the healthcare system.
Due to a large number of people sharing information on social media platforms, an increasing number of efforts focus on social media data to carry out effective ADR detection. 
Despite having achieved impressive performance, the existing methods of ADR detection still suffer from three main challenges. Firstly, researchers have consistently ignored the interaction between domain keywords and other words in the sentence. Secondly, social media datasets suffer from the challenges of low annotated data. Thirdly, the issue of sample imbalance is commonly observed in social media datasets.
To solve these challenges, we propose the \underline{K}nowledge \underline{E}nhanced \underline{S}hallow and \underline{D}eep \underline{T}ransformer(KESDT) model for ADR detection. Specifically, to cope with the first issue, we incorporate the domain keywords into the Transformer model through a shallow fusion manner, which enables the model to fully exploit the interactive relationships between domain keywords and other words in the sentence. To overcome the low annotated data, we integrate the synonym sets into the Transformer model through a deep fusion manner, which expands the size of the samples. To mitigate the impact of sample imbalance, we replace the standard cross entropy loss function with the focal loss function for effective model training.
We conduct extensive experiments on three public datasets including TwiMed, Twitter, and CADEC. The proposed KESDT outperforms state-of-the-art baselines on F1 values, with relative improvements of 4.87\%, 47.83\%, and 5.73\% respectively, which demonstrates the effectiveness of our proposed KESDT. 

\end{abstract}

\section{Introduction}
Adverse drug reactions (ADRs) refer to the harmful and unintended effects that occur after using a medication, which is different from the expected therapeutic results \cite{baber1994international}. These adverse reactions may occur during or after the use of the medication, and their severity can range from mild discomfort to severe and even fatal outcomes. Therefore, timely and accurate detection of potential ADRs is crucial to ensure the safety and effectiveness of medications.

With the rapid development of social media, a large number of users choose to share their medication experiences on social media platforms, leading to an increasing number of researchers conducting ADR detection studies on social media datasets \cite{kanchan2023social}. Although researchers have made valuable contributions to the field of ADR detection, there are still several issues that need to be addressed. 

\begin{figure}[h]
\centering
\vspace{-0.2in}
\includegraphics[width=\textwidth]{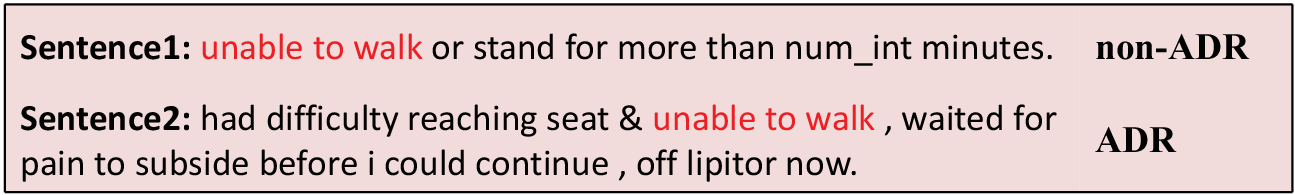}
\vspace{-0.28in}
\caption{An example from the social medial dataset(CADEC).} 
\vspace{-0.22in}
\label{intro}
\end{figure}

Firstly, previous works have overlooked the interaction between domain keywords and other words in the sentence. For example, as shown in Figure 1, although both sentence 1 and sentence 2 contain the adverse reaction term 'unable to walk,' only sentence 2 is labeled as an ADR. This is because sentence 1 does not indicate that the adverse reaction occurred after taking medication, while sentence 2 states that the adverse reaction was caused by taking the drug Lipitor. Therefore, we can conclude that reinforcing the interaction between domain keywords and other words in the sentence can effectively improve the performance of the ADR detection task.

Secondly, social media datasets suffer from the issue of low annotated data. Previous studies have attempted to mitigate this issue by introducing adversarial training \cite{zhang2021adversarial}, adding annotated social media datasets \cite{sarker2015portable}, or using multi-task learning methods \cite{yadav2019unified,chowdhury2018multi}, but these methods inevitably introduce noise and additional annotation workloads. 

Thirdly, the majority of data in social media is unrelated to ADRs, resulting in a severe sample imbalance issue. While literature \cite{huang2022predicting} proposed a weighted cross entropy loss function to address the sample imbalance problem, this loss function only considers the ratio of positive to negative samples and does not take into account the importance of focusing on difficult instances.

To address the aforementioned issues, we propose a novel framework, Knowledge Enhanced Shallow and Deep Transformer (KESDT), for ADR detection. 
Specifically, to cope with the first issue, we incorporate the domain keywords into the Transformer model through a shallow fusion manner, which enables the model to fully exploit the interactive relationships between domain keywords and other words in the sentence. To overcome the low annotated data, we integrate the synonym sets into the Transformer model through a deep fusion manner, which expands the size of the samples. To mitigate the impact of sample imbalance, we replace the standard cross entropy loss function with the focal loss function \cite{lin2017focal,aljohani2023novel} for effective model training.

\begin{itemize}
    \item We propose a new external knowledge integration strategy, which integrates external knowledge into the Transformer via shallow and deep fusion manner respectively. Furthermore, the deep fusion method can be viewed as a novel data augmentation technique. 
    \item We first propose to introduce the focal loss function to solve the sample imbalanced problem in the field of ADRs. 
    \item The results of our experiments indicate that our proposed model exhibits better generalization ability and achieves excellent performance even on small-scale and imbalanced datasets.  
\end{itemize}

\section{Related Work}

\subsection{ADR detection}
The study of ADR detection is a long-standing research problem in the field of bioinformatics.
Early research efforts focused on the detection of ADR from biomedical texts and clinical reports using rule matching\cite{kuhn2010side,benton2011identifying,yates2013adrtrace}. Rule matching methods have made some research progress, but these methods rely on knowledge bases or lexicons, which will lead to limitations in the generalization ability of the models.
With the increase in annotated datasets, a large number of supervised machine-learning methods \cite{bian2012towards,patki2014mining,rastegar2016detecting} have been used to detect ADR. Although machine learning models have greatly improved the detection performance of ADR, these methods rely heavily on domain knowledge and hand-crafted features that are difficult to adapt to new datasets.

In recent years, with the development and application of deep learning methods\cite{zhang2021dual,zhang2022price,zhang2022dynamic}, many deep learning approaches \cite{zhang2021adversarial,huynh2016adverse,alimova2018interactive,wu2018detecting,raval2021exploring,li2020exploiting} have been applied to ADR detection tasks. For example, Huynh et al. \cite{huynh2016adverse} proposed two new convolutional neural network models (CRNN, CNNA) for ADR detection. Alimova et al. \cite{alimova2018interactive} investigated the applicability of an interactive attention network (IAN) in identifying drug adverse reactions from user comments. Wu et al. \cite{wu2018detecting} developed a method with multi-head self-attention and hierarchical tweet representations to detect ADR. In addition, graph neural network methods \cite{wu2023graph,kwak2020drug,shen2021gar,gao2023contextualized} have also been used for ADR detection tasks. For instance, CGEM \cite{gao2023contextualized} combined pre-trained language models with graph neural networks. These studies demonstrate the continuity and urgency of ADR detection research, with many researchers having made significant contributions to the ADR detection task. However, challenges still remain in areas such as semantic interactions, low annotated data, and sample imbalance, posing certain difficulties.

\section{Methodology}
Fig.2 illustrates an overview of the proposed KESDT model structure. KESDT model consists of three main components: (1) \textbf{Shallow fusion layer}: It integrates domain keywords into the Transformer model, which enables the model to fully exploit the interactive relationships between domain keywords and other words in the sentence. (2) \textbf{Deep fusion layer}: It incorporates the synonym sets into the Transformer model, which effectively alleviates the challenges associated with low annotated data. (3) \textbf{Model training}: We introduce the focal loss function to optimize the model.

\begin{figure}[h]
\centering
\vspace{-0.2in}
\includegraphics[width=\textwidth]{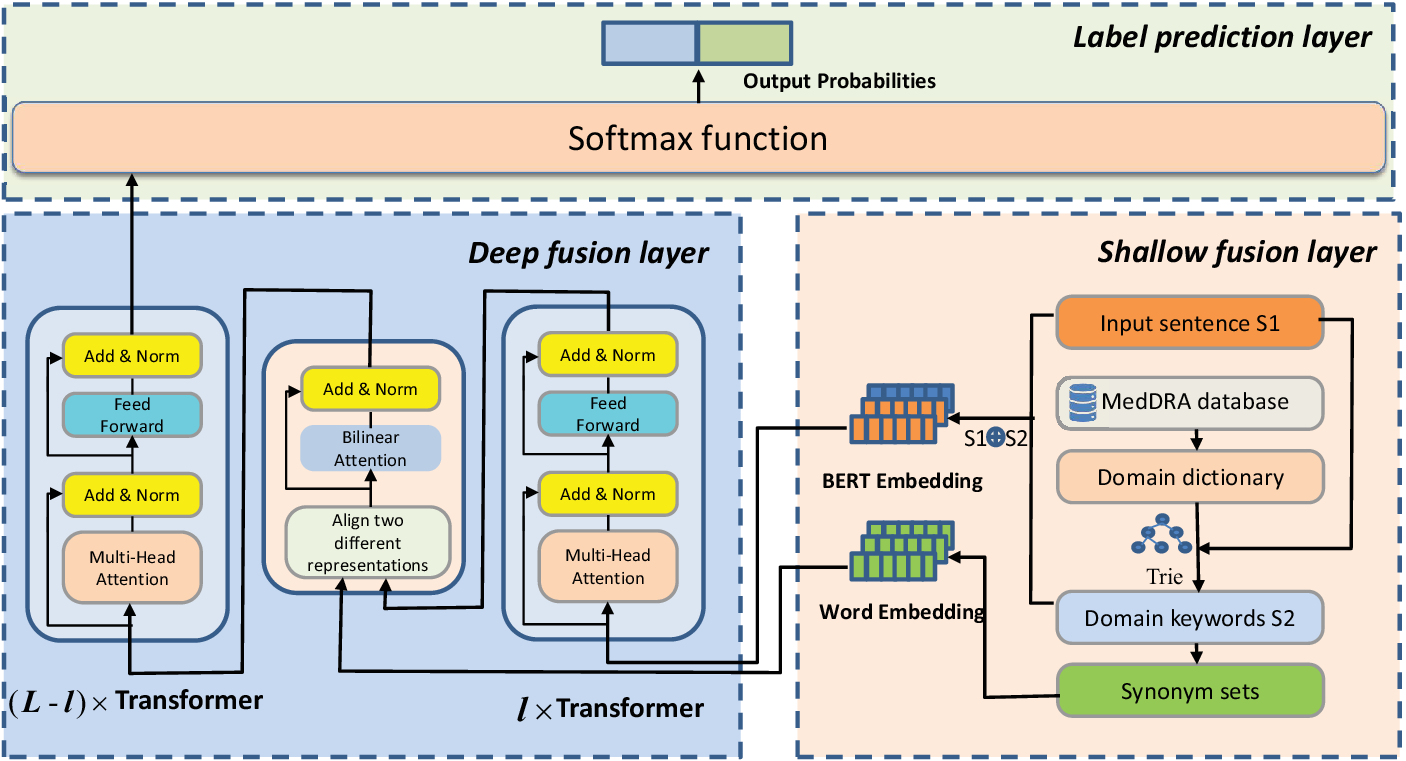}
\vspace{-0.2in}
\caption{The overall framework of KESDT.} 
\vspace{-0.2in}
\label{model}
\end{figure}

\vspace{-0.3in}
\subsection{Problem definition}
In this paper, the ADR detection task is defined as a textual binary classification task, given a social media text sequence ${{S}_{1}}=\{{{w}_{1}},{{w}_{2}},...,{{w}_{n}}\}$, where ${{w}_{i}}$ denotes the $i$-th word in the text and $n$ denotes the length of the sentence, the label $y\in \{0,1\}$ indicates whether the text contains information about ADRs.

\vspace{-0.1in}
\subsection{Shallow fusion layer}
\vspace{-0.1in}
In order to integrate external knowledge into the Transformer in a shallow fusion manner, we propose to construct domain keywords based on external knowledge. There are three sub-steps in the process of integrating domain keywords into the Transformer.

The first step is the construction of a domain dictionary based on the database MedDRA \cite{mozzicato2009meddra}, which contains about 1430 drugs and their known side effects. Specifically, the original database contains 95912 data, we cut the phrases into words by spaces, then pre-processed the data, such as removing special characters like stop words, numbers, etc. In order to ensure the uniqueness of the domain keywords, we specifically removed the words with a length of less than 3. Finally, we obtained a domain dictionary containing 16339 words, which mainly contains some words related to ADRs, such as haematuria, desquamation, hyperalgesia, etc.

The second step is the construction of the domain keywords based on the domain dictionary, Specifically, we first build a Trie based on the domain dictionary, then traverse all the character subsequences of the original text and match them with the Trie to obtain all potential words. We name the potential words as domain keywords and denote them as ${{S}_{2}}=\{{{k}_{1}},{{k}_{2}},...,{{k}_{m}}\}$, where ${{k}_{i}}$ denotes the $i$-th keyword and $m$ is the number of keywords.

Finally, we use the special token ($[\rm{SEP}]$) to concatenate the original text sequence and domain keywords. For a given input sequence 
$S=\{[\rm{CLS}],{{S}_{1}},[\rm{SEP}],{{S}_{2}},[\rm{SEP}]\}$, its input representation $E=\{{{e}_{1}},{{e}_{2}},...,{{e}_{n+m}}\}$ is constructed by summing the corresponding token, segment, and position embeddings. We input $E$ into Transformer encoders of the pre-trained language model, as for pre-trained language models, we use the following three types: they are bert-base-uncased$\footnote{https://huggingface.co/bert-base-uncased}$(\textbf{bert-uncased}), bert-base-cased$\footnote{https://huggingface.co/bert-base-cased}$(\textbf{bert-cased}), biobert-base-cased$\footnote{https://huggingface.co/dmis-lab/biobert-base-cased-v1.2}$(\textbf{biobert})) respectively. each layer of the Transformer can be represented as follows:
\begin{equation}
\label{eq1} 
\begin{array}{lcl}
G={\rm{LN}}(X^{l-1}+{\rm{MHA}}(X^{l-1})), \\[3mm]
X^l={\rm{LN}}(G+{\rm{FFN}}(G))
\end{array}
\end{equation}
where ${{X}^{l}}=\{x_{1}^{l},x_{2}^{l},...,x_{n+m}^{l}\}$ is the $l$-th layer Transformer's output and we set $X^0$ as $E$; ${\rm LN}$ demotes layer normalization; ${\rm MHA}$ denotes the multi-head attention block; ${\rm FFN}$ denotes a feed-forward network.

\subsection{Deep fusion layer}
The deep fusion layer aims to expand the size of the samples. Inspired by the research of Liu et al. \cite{liu2021lexicon}, it contains four sub-steps.

The first step is the construction of synonym sets based on the domain keywords. Specifically, the synonym sets are obtained by word embedding, such as word2vec$\footnote{https://github.com/mmihaltz/word2vec-GoogleNews-vectors}$, which is pre-trained on a large corpus and contains rich semantic information of words. Then the keyword-synonym pair sequence denoted as $\{({{k}_{1}},{{t}_{1}}),({{k}_{2}},{{t}_{2}}),...,({{k}_{m}},{{t}_{m}})\}$, where ${{t}_{i}}=\{{{t}_{i1}},{{t}_{i2}},...,{{t}_{ih}}\}$ denotes the $i$-th synonym sets, $h$ denotes the numbers of synonyms, the $j$-th word in ${{t}_{i}}$ is represented as ${{t}_{ij}}$.

The second step is to align two different representations, we first get the output ${{X}^{l}}=\{x_{1}^{l},x_{2}^{l},...,x_{n+m}^{l}\}$ after $l$ successive Transformer layers, where $x_{i}^{l}\in R$ denotes the $i$-th Character vector if the Character is a domain keyword. Then the synonym sets denote as $x_{i}^{lt}=\{x_{i1}^{lt},x_{i2}^{lt},...,x_{ih}^{lt}\}$, the representation vector $x_{ij}^{lt}$ of $j$-th word matched by the $i$-th Character vector is computed as follows:
\begin{equation}
\label{eq1} 
\begin{array}{lcl}
x_{ij}^{lt}=D({{t}_{ij}})
\end{array}
\end{equation}
where $D$ is a pre-trained word embedding lookup table. We use a linear transformation to align the two different representations:
\begin{equation}
\label{eq1} 
\begin{array}{lcl}
u_{ij}^w=\boldsymbol{W}_1x_{ij}^{lt}+\boldsymbol{b}_1
\end{array}
\end{equation}
where ${\boldsymbol{W}_{1}}$ and ${\boldsymbol{b}_{1}}$ are learnable parameter.

The third step is to pick out the most relevant words from all matched words, we propose to apply a character-to-word attention mechanism. The synonym sets corresponding to each character is denoted as $u_i=(u_{i1}^{lt},...,u_{ih}^{lt})$, where $h$ denotes the total number of the synonymous. The relevancy of each word can be calculated using the attention mechanism and expressed in the following formula:
\begin{equation}
\label{eq1} 
\begin{array}{lcl}
\boldsymbol{r}_i={\rm softmax}(x_i^l\boldsymbol{W}_{2}{u}_i^{\rm{T}})
\end{array}
\end{equation}
where ${\boldsymbol{W}_{2}}$ denotes the learnable parameter determining the importance of synonyms. We represent the sets of synonyms for a word via linear sum as follows:
\begin{equation}
\label{eq1} 
\begin{array}{lcl}
x_i^{lt}=\sum_{j=1}^{h}r_{ij}u_{ij}^{lt}
\end{array}
\end{equation}
Finally, we can enrich the semantics of a word by:
\begin{equation}
\label{eq1} 
\begin{array}{lcl}
\tilde{x}_i^l=x_i^l+x_i^{lt}
\end{array}
\end{equation}
Since there are $L=12$ Transformer layers in the pre-trained language model, we input ${{\tilde{X}}^{l}}=\{\tilde{x}_{1}^{l},\tilde{x}_{2}^{l},...,\tilde{x}_{n+m}^{l}\}$ to the remaining $(L-l)$ Transformers. Then, we can get the hidden outputs representation ${{\tilde{X}}^{L}}=\{\tilde{x}_{1}^{L},\tilde{x}_{2}^{L},...,\tilde{x}_{n+m}^{L}\}$ of the last Transformer layer. Lastly, we apply the fully connected layers and softmax activation functions over the representation of the hidden output ${{\tilde{X}}^{L}}$ and obtain the probability of each class as:
\begin{equation}
\label{eq1} 
\begin{array}{lcl}
{{p}_{i}}=\rm{softmax}(\boldsymbol{W}_{class}\tilde{x}_{i}^{L}+{\boldsymbol{b}_{class}})
\end{array}
\end{equation}
where ${\boldsymbol{W}_{\rm class}}$ and ${\boldsymbol{b}_{\rm class}}$ are learnable parameters.

\subsection{Model training}
In this work, the ADR detection task has a severe label distribution imbalance problem. The standard binary cross entropy loss function is slow to iterate and may deviate from the correct optimization direction. So we introduce the focal loss function to optimize the model. The focal loss function is a variant of the binary cross entropy loss function, which reduces the weight of the contribution of simple samples and allows the model to learn more difficult samples. The focal loss is defined as:
\begin{equation}
\label{eq1} 
\begin{array}{lcl}
{{L}_{Focal}}=-{{(1-p)}^{\gamma }}y\log p
\end{array}
\end{equation}
where $p$ is the predicted output of the network activation function, and $y$ is the true label of the sample. $\gamma \ge 0$ is the modulation factor, which is used to reduce the weight of simple samples and make the model focus more on difficult samples. The focal loss function can attenuate the dominant influence of easy samples on gradient updates, thereby preventing the network from learning a significant amount of irrelevant information. It can also avoid the model to be biased towards the categories with more samples and alleviate the category imbalance problem.

\section{Experiment}
\subsection{Dataset and Evaluation}
To evaluate our proposed model fairly and effectively, we performed 5-fold cross-validation on three publicly available social media datasets. The dataset used in our study is similar to that used by Zhang et al. \cite{zhang2021adversarial} and Li et al. \cite{li2020exploiting}, The details of the datasets are shown in Table 1.

\begin{table}
	\centering
    \begin{tabular}{p{2cm}<{\centering}p{1.5cm}<{\centering}p{1.5cm}<{\centering}p{1.5cm}<{\centering}p{1.5cm}<{\centering}p{2cm}<{\centering}p{1.3cm}<{\centering}}
    \hline
    \textbf{Datasets} & \textbf{Positive} & \textbf{Negative} & \textbf{Total} & \textbf{Max sentence length} & \textbf{Experimental data length} & \textbf{Ratios  of samples}\\
    \hline
    Twitter & 744 & 5727 & 6471 & 46 & 46 & {1:8} \\
    TwitMed & 426 & 1182 & 1608 & 137 & 65 & {1:3} \\
    CADEC & 2478 & 4996 & 7474 & 241 & 70 & {1:2} \\
    \hline
    \end{tabular}
    \vspace{0.05in}
    \caption{Statistical information of social media ADR datasets.}
    \vspace{-0.3in}
    \label{results2}
\end{table}

During the data preprocessing stage, we removed stop words, punctuation, and numbers. Additionally, we used the tweet-preprocessor Python package$\footnote{https://pypi.org/project/tweet-preprocessor/}$ to eliminate URLs, emojis, and reserved words in tweets. 
(1) \textbf{TwiMed} \cite{alvaro2017twimed}: The dataset is composed of two parts: TwiMed-Twitter collected from the social media platform(Twitter), and TwiMed-PubMed collected from biomedical literature. Each document is annotated with disease, symptom, drug, and their relationships. There are three types of relationships: Outcome-negative, Outcome-positive, and Reason-to-use. When the relationship type is Outcome-positive, we label it as ADR. (2) \textbf{CADEC} \cite{karimi2015cadec}: The dataset is collected from medical forums, where each document is labeled with drugs, side effects, symptoms, and diseases. (3) \textbf{Twitter} \cite{sarker2016social}: This dataset is collected from the social media platform Twitter, and each sentence is marked as ADR or non-ADR.

We utilized three metrics, namely precision (P), recall (R), and micro F1-score (F1), to evaluate the performance of the proposed model.

\subsection{Experimental Settings and Baselines}
In our experiments, we used 300-dimensional pre-trained word embeddings from word2vec to represent synonym sets and integrated them between the first and second layers of the pre-trained language model. During training, we fine-tuned both the pre-trained language model and the pre-trained word embeddings, with each domain keyword matching up to five synonyms. We used Adam as the optimizer and trained the model for 15 epochs with a batch size of 64 for all datasets, with a learning rate of 1e-5 for CADEC, 3e-5 for Twitter, and 5e-5 for TwiMed. Additionally, we set the dropout value to 0.1 and the modulation factor $\gamma$ of focal loss to 2. All experiments were implemented in Python 3.7 and PyTorch 1.10 framework and trained on NVidia TITAN XP GPU.

The ADR detection models compared include the following methods: (1) \textbf{RCNN} \cite{huynh2016adverse}: a recurrent convolutional neural network model. (2) \textbf{HTR-MSA} \cite{wu2018detecting}: a model that combines multi-head self-attention and hierarchical tweet representation. (3) \textbf{CNN+corpus} \cite{li2020exploiting}: a model that adds additional annotated corpus to the CNN method. (4) \textbf{cnn+transfer} \cite{li2020exploiting}: a model that combines cnn and transfer learning. (5) \textbf{ATL} \cite{li2020exploiting}: a model that combines adversarial training and transfer learning. (6) \textbf{ANNSA} \cite{zhang2021adversarial}: a model that combines sentiment attention mechanism and adversarial learning. (7) \textbf{Baseline(bert-uncased)}: a model that combines bert-uncased pre-trained language model and fully connected layer.
 
\vspace{-0.2in}
\subsection{Results and Discussions}
Table 2, Table 3, and Table 4 present the experimental results of KESDT and the compared models on three datasets (CADEC, Twitter, and TwiMed). Bolded text indicates the best results. 

\vspace{-0.1in}
\begin{table}
    \centering
		\centering
        \begin{tabular}{p{4cm}<{\centering}p{2cm}<{\centering}p{2cm}<{\centering}p{2cm}<{\centering}} 
        \hline
        \textbf{Model} & \textbf{P(\%)} & \textbf{R(\%)} & \textbf{F1(\%)} \\
        \hline
        \tabincell{c}{RCNN \cite{huynh2016adverse}} &81.99 &76.63 &79.22 \\
        \tabincell{c}{HTR+MSA \cite{wu2018detecting}} &81.77 &77.64 &79.65 \\
        \tabincell{c}{CNN+corpus \cite{li2020exploiting}} &85.40 &75.99 &80.42 \\
        \tabincell{c}{CNN+transfer \cite{li2020exploiting}} &84.75 &79.38 &81.98 \\
        \tabincell{c}{ATL \cite{li2020exploiting}} &84.30 &81.28 &82.76 \\
        \tabincell{c}{ANNSA \cite{zhang2021adversarial}} &82.73 &83.52 &83.06 \\        
        \hline
        \tabincell{c}{Baseline(bert-uncased)} &87.06 &86.54 &86.73 \\
        \tabincell{c}{KESDT(bert-uncased)} &\textbf{88.16} &\textbf{87.63} &\textbf{87.82} \\
        \tabincell{c}{KESDT(bert-cased)} &87.84 &87.43 &87.52 \\
        \tabincell{c}{KESDT(biobert)} &87.41 &87.50 &87.42 \\
        \hline
        \end{tabular}
        \vspace{0.05in}
        \caption{Experimental results on the CADEC dataset.}
        \vspace{-0.25in}
        \label{results1}
\end{table}

\vspace{-0.2in}
\begin{table}
    \centering
		\centering
        \begin{tabular}{p{4cm}<{\centering}p{2cm}<{\centering}p{2cm}<{\centering}p{2cm}<{\centering}} 
        \hline
        \textbf{Model} & \textbf{P(\%)} & \textbf{R(\%)} & \textbf{F1(\%)} \\
        \hline
        \tabincell{c}{RCNN \cite{huynh2016adverse}} &50.00 &42.88 &46.17 \\
        \tabincell{c}{HTR+MSA \cite{wu2018detecting}} &37.06 &58.33 &45.33 \\
        \tabincell{c}{CNN+corpus \cite{li2020exploiting}} &47.94 &43.82 &45.79 \\
        \tabincell{c}{CNN+transfer \cite{li2020exploiting}} &60.23 &35.62 &44.76 \\
        \tabincell{c}{ATL \cite{li2020exploiting}} &56.26 &39.25 &46.24 \\
        \tabincell{c}{ANNSA \cite{zhang2021adversarial}} &49.10 &50.46 &48.84 \\        
        \hline
        \tabincell{c}{Baseline(bert-uncased)} &44.26 &50.00 &46.96 \\
        \tabincell{c}{KESDT(bert-uncased)} &70.40 &\textbf{75.58} &\textbf{72.20} \\
        \tabincell{c}{KESDT(bert-cased)} &73.13 &70.27 &71.24 \\
        \tabincell{c}{KESDT(biobert)} &\textbf{73.56} &69.86 &71.29 \\
        \hline
        \end{tabular}
        \vspace{0.05in}
        \caption{Experimental results on the Twitter dataset.}
        \vspace{-0.25in}
        \label{results1}
\end{table}

\vspace{-0.2in}
\begin{table}
    \centering
		\centering
        \begin{tabular}{p{4cm}<{\centering}p{2cm}<{\centering}p{2cm}<{\centering}p{2cm}<{\centering}} 
        \hline
        \textbf{Model} & \textbf{P(\%)} & \textbf{R(\%)} & \textbf{F1(\%)} \\
        \hline
        \tabincell{c}{RCNN \cite{huynh2016adverse}} &68.52 &66.43 &67.46 \\
        \tabincell{c}{HTR+MSA \cite{wu2018detecting}} &66.58 &63.62 &65.07 \\
        \tabincell{c}{CNN+corpus \cite{li2020exploiting}} &60.51 &61.50 &61.00 \\
        \tabincell{c}{CNN+transfer \cite{li2020exploiting}} &69.58 &61.74 &65.42 \\
        \tabincell{c}{ATL \cite{li2020exploiting}} &70.84 &65.02 &67.81 \\      
        \hline
        \tabincell{c}{Baseline(bert-uncased)} &64.58 &62.42 &60.35 \\
        \tabincell{c}{KESDT(bert-uncased)} &71.22 &69.15 &68.63 \\
        \tabincell{c}{KESDT(bert-cased)} &\textbf{71.72} &\textbf{72.13} &\textbf{71.11} \\
        \tabincell{c}{KESDT(biobert)} &69.69 &69.53 &69.57 \\
        \hline
        \end{tabular}
        \vspace{0.05in}
        \caption{Experimental results on the TwiMed dataset.}
        \vspace{-0.25in}
        \label{results1}
\end{table}

(1) From the experimental results, it can be seen that our proposed KESDT framework performs significantly better than the compared models in the task of ADR detection. Previous works (CNN+corpus, CNN+Transfer, ATL) have attempted to address the issue of limited data by incorporating additional annotated datasets or employing transfer learning techniques to enhance ADR detection performance. However, these methods inevitably introduce noise and require additional annotation efforts. Our KESDT model enhances the sample size by introducing a new data augmentation method, which effectively improves the model's performance.

(2) We investigated the impact of different pre-trained language models (bert-uncased, bert-cased, and biobert) on the performance of the KESDT model. Based on these results, we found that the KESDT model has good generalization performance, not only performing well on bert-uncased but also achieving good results on other pre-trained language models, such as bert-cased and biobert.

\vspace{-0.1in}
\subsection{Ablation experiments}
Table 5, Table 6, and Table 7 present the ablation experiment results of the KESDT model on three different datasets. All ablation experiments were based on the bert-uncased pre-trained language model.

\vspace{-0.1in}
\begin{table}
    \centering
		\centering
        \begin{tabular}{p{4cm}<{\centering}p{2cm}<{\centering}p{2cm}<{\centering}p{2cm}<{\centering}p{2cm}<{\centering}} 
        \hline
        \textbf{Model} & \textbf{P(\%)} & \textbf{R(\%)} & \textbf{F1(\%)}  &   \textbf{$\Delta$F1(\%)} \\
        \hline
        KESDT & 88.16 & \textbf{87.63} & \textbf{87.82} & - \\
        KESDT-keywords & \textbf{88.21} & 87.24 & 87.56 & -0.26 \\
        KESDT-synonyms & 87.84 & 87.06 & 87.30 & -0.52 \\
        KESDT-(focal loss) & 87.41 & 87.25 & 87.21 & -0.61 \\
        \tabincell{c}{KESDT-keywords-synonyms} &86.95 &86.04 &86.43 &-1.39\\
        Baseline(bert-uncased) & 87.06 & 86.54 & 86.73 & -1.09 \\
        \hline
        \end{tabular}
        \vspace{0.05in}
        \caption{Ablation experiments on the CADEC dataset.}
        \vspace{-0.25in}
        \label{results1}
\end{table}

\begin{table}
    \centering
		\centering
        \begin{tabular}{p{4cm}<{\centering}p{2cm}<{\centering}p{2cm}<{\centering}p{2cm}<{\centering}p{2cm}<{\centering}} 
        \hline
        \textbf{Model} & \textbf{P(\%)} & \textbf{R(\%)} & \textbf{F1(\%)}  &   \textbf{$\Delta$F1(\%)} \\
        \hline
        KESDT & 71.22 & \textbf{69.15} & \textbf{68.63} & - \\
        KESDT-keywords & 70.26 & 68.53 & 68.37 & -0.26 \\
        KESDT-synonyms & \textbf{72.59} & 64.79 & 63.70 & -4.93 \\
        KESDT-(focal loss) & 70.30 & 68.53 & 66.68 & -1.95 \\
        \tabincell{c}{KESDT-keywords-synonyms} &66.17 &66.15 &65.50 &-3.13\\
        Baseline(bert-uncased) & 64.58 & 62.42 & 60.35 & -8.28 \\
        \hline
        \end{tabular}
        \vspace{0.05in}
        \caption{Ablation experiments on the TwiMed dataset.}
        \vspace{-0.25in}
        \label{results1}
\end{table}

\begin{table}
    \centering
		\centering
        \begin{tabular}{p{4cm}<{\centering}p{2cm}<{\centering}p{2cm}<{\centering}p{2cm}<{\centering}p{2cm}<{\centering}} 
        \hline
        \textbf{Model} & \textbf{P(\%)} & \textbf{R(\%)} & \textbf{F1(\%)}  &   \textbf{$\Delta$F1(\%)} \\
        \hline
        KESDT & 70.40 & \textbf{75.58} & \textbf{72.20} & - \\
        KESDT-keywords & 71.69 & 73.06 & 72.00 & -0.20 \\
        KESDT-synonyms & \textbf{73.09} & 71.55 & 71.62 & -0.58 \\
        KESDT-(focal loss) & 44.26 & 50.00 & 46.96 & -25.24 \\
        \tabincell{c}{KESDT-keywords-synonyms} &73.08 &69.51 &70.80 &-1.40\\
        Baseline(bert-uncased) & 44.26 & 50.00 & 46.96 & -25.24 \\
        \hline
        \end{tabular}
        \vspace{0.05in}
        \caption{Ablation experiments on the Twitter dataset.}
        \vspace{-0.25in}
        \label{results1}
\end{table}

\vspace{-0.1in}
(1) To evaluate the impact of incorporating domain keywords, we conducted ablation experiments by removing the domain keywords module from the KESDT model. Compared to the KESDT model, the KESDT-keywords model showed a decrease in F1 score by 0.26\%, 0.26\%, and 0.2\% on the CADEC, TwiMed, and Twitter datasets, respectively. These results confirm the importance of domain keywords information for the ADR detection task.

(2) To evaluate the enhancement effect of introducing synonym sets, we removed the synonym sets part. After removing the synonym sets part, the F1 values of the KESDT-synonyms model on the three datasets (CADEC, TwiMed, and Twitter) decreased by 0.52\%, 4.93\%, and 0.58\%, respectively. Especially on the small-scale TwiMed dataset, the decrease was the most significant, indicating that introducing synonym sets as a data augmentation method can significantly improve the model performance.

(3) To explore the impact of the focal loss function on imbalanced datasets, we replaced the focal loss with the standard cross entropy loss. We found that the model achieved the most significant improvement on the imbalanced dataset(Twitter). The F1 score of the KESDT-(focal loss) model decreased by 25.24\% compared to the KESDT model, indicating the importance of the focal loss function in addressing sample imbalance issues.

(4) To evaluate the impact of the interaction between domain keywords and synonym sets on model performance, we removed both the domain keywords module and the synonym sets module simultaneously. We observed that the KESDT-keywords-synonyms model performed worse in terms of F1 score compared to the KESDT-keywords or KESDT-synonyms models. This indicates the importance of simultaneously incorporating domain keywords and synonym sets for improving model performance.

(5) Finally, we evaluated the combined effect of introducing domain keywords, synonym sets, and replacing standard cross entropy loss with focal loss. We found that the performance of the baseline(bert-uncased) model on three datasets significantly decreased, further confirming the effectiveness of each module we introduced.
\vspace{-0.1in}

\section{Conclusion}
\vspace{-0.1in}
In the field of biomedicine, the detection of ADR represents a meaningful and fundamental task. To address the current challenges in ADR detection, we propose a novel neural approach called Knowledge Enhanced Shallow and Deep Transformer (KESDT). In future research, we will explore two directions: (1) In-context learning has been widely applied in natural language processing, and we aim to investigate its potential to reduce the dependence of model training on annotated data. (2) The experiments have demonstrated the importance of domain keywords and synonym sets for this task, and we will design a more effective method to select them. 

\section*{Acknowledgement}
This work is partially supported by grant from the Natural Science Foundation of China (No.62076046, No.62006130), Inner Monoglia Science Foundation (No.2022MS06028). This work is also supported by the National \& Local Joint Engineering Research Center of Intelligent Information Processing Technology for Mongolian and the Inner Mongolia Directly College and University Scientific Basic in 2022.

\bibliographystyle{splncs04}
\bibliography{references}

\end{document}